# Study of Efficient Technique Based On 2D Tsallis Entropy For Image Thresholding

Mohamed. A. El-Sayed[a,c], S. Abdel-Khalek[b,d], and Eman Abdel-Aziz[b]

[a] Mathematics department, Faculty of Science, Fayoum University, 63514 Fayoum, Egypt
[b] Mathematics department, Faculty of Science, Sohag University, 82524 Sohag, Egypt
[c] CS department, Faculty of Computers and Information Science, Taif Univesity, 21974 Taif, KSA
[d] Mathematics department, Faculty of Science, Taif Univesity, 21974 Taif, KSA
drmasayed@yahoo.com

**Abstract:** Thresholding is an important task in image processing. It is a main tool in pattern recognition, image segmentation, edge detection and scene analysis. In this paper, we present a new thresholding technique based on two-dimensional Tsallis entropy. The two-dimensional Tsallis entropy was obtained from the two-dimensional histogram which was determined by using the gray value of the pixels and the local average gray value of the pixels, the work it was applied a generalized entropy formalism that represents a recent development in statistical mechanics. The effectiveness of the proposed method is demonstrated by using examples from the real-world and synthetic images. The performance evaluation of the proposed technique in terms of the quality of the thresholded images are presented. Experimental results demonstrate that the proposed method achieve better result than the Shannon method.

## 1. INTRODUCTION

The principal assumption of the use of global thresholding as a segmentation technique is that "objects" and "backgrounds" can be distinguished by inspecting only image gray level values. Segmentation consist in subdividing an image into its constituent part and extracting those of interest. Many techniques for global thresholding have been developed over the years to segment images and recognize patterns (e.g. [6, 8, 9, 10, 11, 12, 13]).

Threshold selection methods can be classified into two groups, namely, global methods and local methods. A global thresholding technique thresholds the entire image with a single threshold value obtained by using the gray level histogram (which is an approximation of the gray level probability density function) of the image. Local thresholding methods partition the given image into a number of subimages and determine a threshold for each of the subimages. Global thresholding methods are easy to implement and are computationally less involved. As such they serve as popular tools in a variety of image processing applications such as segmentation of synthetic radar images [19], video indexing through scene cut [18], biomedical image analysis [4, 17], image sequence segmentation [5], and text enhancement [7]. For a survey on threshold selection techniques the reader is encouraged to see Refs. [4, 13].

In Ref. [1], Abutaleb extended the entropy based thresholding method of Kapur et al [6], using the two-dimensional entropy. The two-dimensional entropies were obtained from a two-dimensional histogram which was determined by using the gray value of the pixels and the local average gray value of the pixels. This extension due to Abutaleb was refined by Brink [2] by maximizing the smaller of two entropies instead of maximizing the sum of the two entropies of the background class and object class.

Many references extended one-dimensional entropic thresholding approach to two-dimensional entropic approach on the basis of two-dimensional histogram. Recently Sahoo and Arora [12] proposed a two-dimensional Renyi thresholding method [14] which performed better than one-dimensional method when images are corrupted with noise.

While the extension of one-dimensional approaches to two-dimensional histogram results in much better segmentation, it gives rise to the exponential increment of computational time. We only consider the entropy-based methods for comparison. The computational complexity of Brink's method [2], using a "maximum" optimization procedure, is bounded by $O(n^4)$, and, after improved by Chen et al. [3] is still bounded by $O(n^{8/3})$, where $n$ is the number of gray levels. But the computational complexity of the method of Kapur et al. [6] is only bounded by $O(n^{3/2})$. In Ref. [16], Wu et al. give a fast recurring two-dimensional thresholding algorithm to improve the computation time of the thresholding method due to Abutaleb [1].

In this paper, we propose an automatic global thresholding technique based on two-dimensional Tsallis's entropy. The two-dimensional Tsallis's entropy was obtained from the two-dimensional histogram which was determined by using the gray value of the pixels and the local average gray value of the pixels. This new method extends a method due to M. Portes de Albuquerque et al [10].

This paper is organized as follows: in Section 2 presents some fundamental concepts of the mathematical setting of the threshold selection and Tsallis entropy. Section 3, we describe the newly proposed thresholding method. In Section 4, we report the effectiveness of our thresholding method when applied to some real-world and synthetic images. In Section 5, we present some concluding remarks about our method.

## 2. PRELIMINARIES

The entropy of a discrete source is often obtained from the probability distribution $p=\{p_i\}$. Therefore, $0 \leq p_i \leq 1$ and $\sum_{i=1}^{k} p_i = 1$, and the Shannon entropy is described as $S = -\sum_{i=1}^{k} p_i ln(p_i)$, being $k$ the total number of states. If we consider that a physical system can be decomposed in two statistical independent subsystems $A$ and $B$, the Shannon entropy has the extensive property (additivity) $S(A+B) = S(A) + S(B)$. This formalism has been shown to be restricted to the Boltzmann-Gibbs-Shannon (BGS) statistics. However, for nonextensive physical systems, some kind of extension appears to become necessary. Tsallis [15] has proposed a generalization of the BGS statistics which is useful for describing the thermostatistical properties of nonextensive systems. It is based on a generalized entropic form, $S_q = \frac{1}{q-1}(1 - \sum_{i=1}^{k}(p_i)^q)$, where the real number $q$ is a entropic index that characterizes the degree of nonextensivity. This expression recovers to BGS entropy in the limit $q \to 1$. Tsallis entropy has a nonextensive property for statistical independent systems, defined by the following pseudo additivity entropic rule

$$S_q(A+B) = S_q(A) + S_q(B) + (1-q).S_q(A).S_q(B).$$

Let $f(m, n)$ be the gray value of the pixel located at the point $(m, n)$. In a digital image $\{f(m,n) \mid m \in \{1,2,...,M\}, n \in \{1,2,...,N\}\}$ of size $M \times N$, let the histogram be $h(x)$ for $x \in \{0,1,2,...,255\}$. For the sake of convenience, we denote the set of all gray levels $\{0,1,2,..., 255\}$ as $G$. Global threshold selection methods usually use the gray level histogram of the image. The optimal threshold is determined by optimizing a suitable criterion function obtained from the gray level distribution of the image and some other features of the image.

Let $t$ be a threshold value and $B = \{b_0, b_1\}$ be a pair of binary gray levels with $\{b_0, b_1\} \in G$. Typically $b_0$ and $b_1$ are taken to be 0 and 255, respectively. The result of thresholding an image function $f(m, n)$ at gray level $t$ is a binary function $f_t(m, n)$ such that $f_t(m,n) = b_0$ if $f_t(m,n) \leq t$ otherwise, $f_t(m,n) = b_1$. In general, a thresholding method determines the value $t^*$ of $t$ based on a certain criterion function. If $t^*$ is determined solely from the gray level of each pixel, the thresholding method is point dependent [13].

In order to compute the two-dimensional histogram of a given image we proceed as follow. Calculate the average gray value of the neighborhood of each pixel. Let $g(x, y)$ be the average of the neighborhood of the pixel located at the point $(x, y)$. The average gray value for the $3 \times 3$ neighborhood of each pixel is calculated as

$$g(x, y) = \left\lfloor \frac{1}{9} \sum_{a=-1}^{1} \sum_{b=-1}^{1} f(x+a, y+b) \right\rfloor,$$

where $\lfloor r \rfloor$ denotes the integer part of the number *r*. While computing the average gray value, disregard the two rows from the top and bottom and two columns from the sides. The pixel's gray value, *f(x, y)*, and the average of its neighborhood, *g(x, y)*, are used to construct a two-dimensional histogram using : $h(k,m) = Prob($ *f(x, y) = k* and *g(x, y) = m* $)$, *where* $\{k, m\} \in G$.

For a given image, there are several methods to estimate this density function. One of the most frequently used methods is the *method of relative frequency*. The normalized histogram $\hat{h}(k,m)$ is approximated by: number of elements in the event ( *f(x, y) = k* and *g(x, y) = m* ), divided by number of elements in the sample space. Hence, $\hat{h}(k,m)$ can be calculate as number of pixels with gray value *k* and average gray value *m*, divided by number of pixels in the image.

The total number of frequencies (occurrences), $n_{(i,j)}$ of the pair *(i, j)*, divided by the total number of pixels, $N \times M$, defines a joint probability mass function, *p(i, j)*. Thus

$$p(i, j) = \frac{n_{(i,j)}}{M \times N} \quad \text{for } i, j = 0, 1, \ldots, 255$$

## 3. THE PROPOSED ALGORITHM:

The threshold is obtained through a vector *(t, s)* where *t*, for *f(x, y)*, represents the threshold of the gray level of the pixel and *s*, for *g(x, y)*, represents the threshold of the average gray level of the pixel's neighborhood. The frequency of occurrence of each pair of gray values is calculated. From this a surface can be drawn that will have two peaks and one valley. The object and background correspond to the peaks and can be separated by selecting the vector *(t, s)* that maximizes the sum of two class entropies. Using this vector *(t, s)*, the histogram is divided into four quadrants (see Fig. 1). We denote the first quadrant by $[t + 1, 255] \times [0, s]$, the second quadrant by $[0, t] \times [0, s]$, the third quadrant by $[0, t] \times [s+1, 255]$, and the fourth quadrant by $[t+1, 255] \times [s + 1, 255]$.

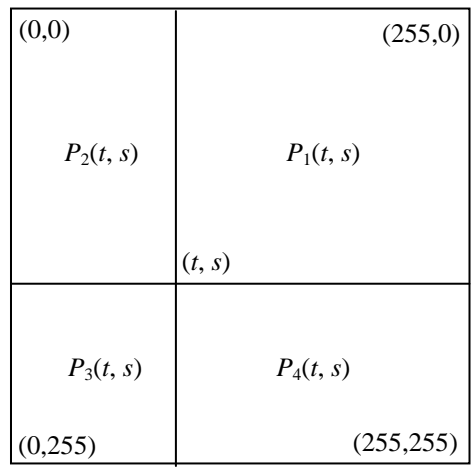

Fig. 1 Quadrants in the 2D histogram due to thresholding at *(t, s)*.

Since two of the quadrants, first and third, contain information about edges and noise alone, they are ignored in the calculation. Because the quadrants which contain the object and the background, second and fourth, are considered to be independent distributions, the probability values in each case must be normalized in order for each of the quadrants to have a total probability equal to 1.
Our normalization is accomplished by using a posteriori class probabilities, $P_2(t, s)$ and $P_4(t, s)$. We assume that the contribution of the quadrants which contains the edges and noise is negligible, hence we further approximate $P_4(t, s)$ as $P_4(t, s) \approx 1 - P_2(t, s)$. The Tsallis entropy of order *q* of an image is defined as:

$$S_q = \frac{1}{q-1}[1 - \sum_{i=0}^{255} \sum_{j=0}^{255} p(i, j)^q ]$$

where $q \neq 1$ is a positive real parameter. Since $\lim_{q \to 1} S_q = S$, Tsallis entropy $S_q$ is a one parameter generalization of the Shannon entropy $S$, where

$$S = -\sum_{i=0}^{255} \sum_{j=0}^{255} p(i,j) \ln p(i,j)$$

Tsallis entropies associated with object and background distributions are given by

$$S_q^A(t,s) = \frac{1}{q-1}[1 - \sum_{i=0}^{t} \sum_{j=0}^{s} (\frac{p(i,j)}{P_2(t,s)})^q]$$

and

$$S_q^B(t,s) = \frac{1}{q-1}[1 - \sum_{i=t+1}^{255} \sum_{j=s+1}^{255} (\frac{p(i,j)}{1-P_2(t,s)})^q]$$

Here we have assumed that the off-diagonal probabilities are negligible and $S_q^B(t,s)$ is computed by using $1 - P_2(t,s)$ instead of $P_4(t,s)$. We try to maximize the information measure between the two classes (object and background). When $S_q(t,s)$ is maximized, the luminance pair $(t,s)$ that maximizes the function is considered to be the optimum threshold pair $(t^*, s^*)$ [6]

$$(t^*(q), s^*(q)) = Arg \max_{(t,s) \in G \times G} S_q^{A+B}(t,s)$$
$$= Arg \max_{(t,s) \in G \times G} [S_q^A(t,s) + S_q^B(t,s) + (1-q).S_q^A(t,s).S_q^B(t,s)].$$

For a priori chosen $q$, we will use only the optimal threshold $t^*(q)$ to threshold an image.

**Theorem 1:** *The threshold value equals to the same value found by Shannon's method when $q \to 1$.*

**Proof:** The limiting case of the proposed extension is Shannon's method. To see this, compute the limiting value of $S_q^A(t,s)$ and $S_q^B(t,s)$ as $q \to 1$. Hence,

$$\lim_{q \to 1} S_q^{A+B}(t,s) = \lim_{q \to 1}[S_q^A(t,s) + S_q^B(t,s) + (1-q).S_q^A(t,s).S_q^B(t,s)].$$
$$= \lim_{q \to 1}[S_q^A(t,s) + S_q^B(t,s)].$$
$$= \lim_{q \to 1}[\frac{1}{q-1}(1 - \sum_{i=0}^{t} \sum_{j=0}^{s} (\frac{p(i,j)}{P_2(t,s)})^q)] + \lim_{q \to 1}[\frac{1}{q-1}(1 - \sum_{i=t1}^{255} \sum_{j=s+1}^{255} (\frac{p(i,j)}{1-P_2(t,s)})^q)]$$
$$= -\lim_{q \to 1} \frac{d}{dq}(\sum_{i=0}^{t} \sum_{j=0}^{s} (\frac{p(i,j)}{P_2(t,s)})^q) - \lim_{q \to 1} \frac{d}{dq}(\sum_{i=t1}^{255} \sum_{j=s+1}^{255} (\frac{p(i,j)}{1-P_2(t,s)})^q)$$
$$= -\lim_{q \to 1} \sum_{i=0}^{t} \sum_{j=0}^{s} \frac{d}{dq}(\frac{p(i,j)}{P_2(t,s)})^q - \lim_{q \to 1} \sum_{i=t1}^{255} \sum_{j=s+1}^{255} \frac{d}{dq}(\frac{p(i,j)}{1-P_2(t,s)})^q$$

but, $\frac{d}{dq}(a^q) = e^{q \ln a}. \ln a$,

i.e. $\lim_{q \to 1} \frac{d}{dq}(a^q) = e^{\ln a}. \ln a = a. \ln a$

hence,

$$\lim_{q \to 1} S_q^{A+B}(t,s) = -\sum_{i=0}^{t} \sum_{j=0}^{s} (\frac{p(i,j)}{P_2(t,s)}) \ln(\frac{p(i,j)}{P_2(t,s)}) - \sum_{i=t1}^{255} \sum_{j=s+1}^{255} (\frac{p(i,j)}{P_2(t,s)}) \ln(\frac{p(i,j)}{P_2(t,s)})$$
$$= S^A(t,s) + S^B(t,s)$$

Therefore, when $q \to 1$, the threshold value equals to the same value found by Shannon's method. Thus this proposed method includes Shannon's method as a special case. The following expression can be used as a criterion function to obtain the optimal threshold at $q \to 1$.

$$(t^*(1), s^*(1)) = Arg \max_{(t,s) \in G \times G} [S^A(t,s) + S^B(t,s)].$$

In order to reduce the execution time, we take $t=s$, i.e. the value of $t$ is lies on the diagonal of quadrants in the 2D histogram and the calculation on only two square matrices, $P_2$ with $t \times t$ and $P_4$ with $(255-t) \times (255-t)$. See Figure 2.

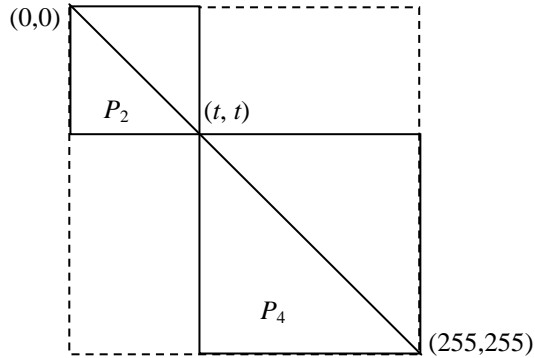

Fig. 2 Quadrants in the 2D histogram due to thresholding at $(t, t)$.

The complete *Tsallis2D* algorithm can now be described as follows:
**Algorithm *Tsallis2D*;**
  **Input:** A digital grayscale image $A$ of size $M \times N$.
  **Output:** The optimal threshold $t^*(q)$ of $A$.
  Begin
    1. Let $f(x, y)$ be the original gray value of the pixel at the point $(x, y)$, $x=1..M$, $y=1..N$.
    2. Calculate the average gray level value $g(x, y)$ in a $3 \times 3$ neighborhood around the pixel $(x, y)$, according to $g(x, y) = \left\lfloor \frac{1}{9} \sum_{a=-1}^{1} \sum_{b=-1}^{1} f(x+a, y+b) \right\rfloor$.

    3. Calculate the joint probability mass function, $p(i, j) = \frac{n_{(i,j)}}{M \times N}$, for $i, j = 0, 1, \ldots, 255$.

    4. If $q \neq 1$ Then
$$(t^*(q), t^*(q)) = Arg \max_{(t,s) \in G \times G} [S_q^A(t,t) + S_q^B(t,t) + (1-q).S_q^A(t,t).S_q^B(t,t)].$$
    Else
$$(t^*(1), t^*(1)) = Arg \max_{(t,s) \in G \times G} [S_q^A(t,t) + S_q^B(t,t)]$$
    EndIf.
    5. For $x=1..M$, $y=1..N$:
       If $f_t(x, y) \leq t^*$ then $f_t(x, y) = 0$ Else $f_t(x, y) = 255$ EndIf.
  End.

## 4. EXPERIMENTAL RESULTS:

In this section, we discuss the experimental results obtained using the proposed method. This discussion includes the choice of the optimal threshold and the presentation of the optimal threshold values of some real-world and synthetic images. These images are: bacteria.tif, blood1.tif, bonemarr.tif, cameraman.tif, cell.tif, eight.tif, flowers.tif, ic.tif, kids.tif, magnetic.tif, moon.tif, mri.tif, rice.tif; root.tif, rose.tif, saturn.tif, shadow.tif, shot1.tif, testpat1.tif, and tire.tif. and they are shown in Figs. 3–20. Our analysis is based on how much information is lost due to thresholding. In this analysis, given two thresholded images of a same original image, we prefer the one which lost the least amount of information. The optimal threshold value was computed by the proposed method for these twenty images. Table 1 lists the optimal threshold values that are found for these images for $q$ values equal to 1.0, 0.3, 0.5, 0.7, 0.9, 1.0 and 2.0, respectively.

The original images together with their histograms and the thresholded images obtained by using the optimal threshold of some values $t^*$ are displayed side by side in Figs. 3–20. Using the above twenty images and also some other images, we conclude that when $q$ value lies between 0 and 1, our proposed method produced good optimal threshold values. Moreover, the optimal threshold value does not change very much when the fractional $q$ value changes a little. However, when $q$ was greater than 1, this proposed method did not produce good threshold values. In fact the threshold values produced were unacceptable (see the last column of Table 1).

When the value of $q$ was one, the threshold value produced was not always a good threshold value (see Figs. 3–20). This new method performs better with fractional values of $q$. In this method of thresholding, we have used in addition to the original gray level function $f(x, y)$, a function $g(x, y)$ that is the average gray level value in a 3×3 neighborhood around the pixel $(x, y)$. This approach can be extended to an image pyramid, where an image on the next higher level is composed of average gray level values computed for disjoint 3×3 squares. From the point of view of computational time and image quality, a neighborhood size of 3 × 3 with $q$ value around 0.1 would be ideal for thresholding with this proposed method.

Table 1 The optimal threshold values $t^*$ for various values of $q$

| Image | $t^*(…)$ | | | | | | |
|---|---|---|---|---|---|---|---|
| | 0.1 | 0.3 | 05. | 0.7 | 0.9 | 1 | 2 |
| bacteria | 102 | 102 | 102 | 102 | 102 | 154 | 6 |
| blood1 | 141 | 141 | 141 | 141 | 141 | 160 | 141 |
| bonemarr | 134 | 149 | 149 | 149 | 157 | 216 | 157 |
| cameraman | 102 | 84 | 84 | 84 | 84 | 119 | 84 |
| cell | 123 | 119 | 119 | 219 | 219 | 113 | 219 |
| eight | 154 | 154 | 154 | 154 | 154 | 228 | 154 |
| flowers | 120 | 120 | 120 | 120 | 120 | 70 | 86 |
| ic | 132 | 150 | 150 | 150 | 150 | 23 | 252 |
| kids | 44 | 44 | 44 | 44 | 44 | 45 | 73 |
| magnetic | 117 | 117 | 117 | 117 | 117 | 151 | 20 |
| moon | 132 | 132 | 85 | 85 | 55 | 197 | 55 |
| mri | 131 | 107 | 96 | 77 | 77 | 101 | 55 |
| rice | 131 | 115 | 115 | 115 | 115 | 96 | 115 |
| medical | 145 | 145 | 160 | 160 | 160 | 132 | 195 |
| rose | 126 | 126 | 109 | 109 | 62 | 40 | 20 |
| saturn | 123 | 119 | 119 | 96 | 45 | 165 | 41 |
| shadow | 131 | 113 | 113 | 113 | 107 | 197 | 107 |
| shot1 | 121 | 165 | 165 | 165 | 165 | 161 | 165 |
| testpat1 | 121 | 118 | 118 | 118 | 104 | 187 | 104 |
| tire | 123 | 116 | 114 | 114 | 114 | 162 | 190 |

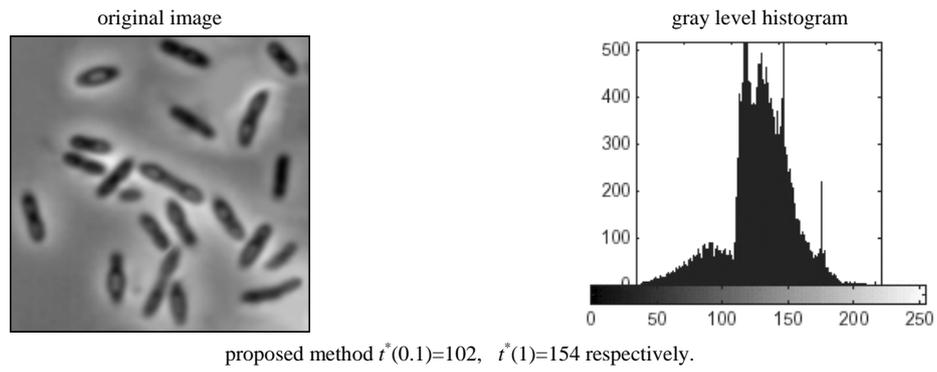

proposed method $t^*(0.1)=102$, $t^*(1)=154$ respectively.

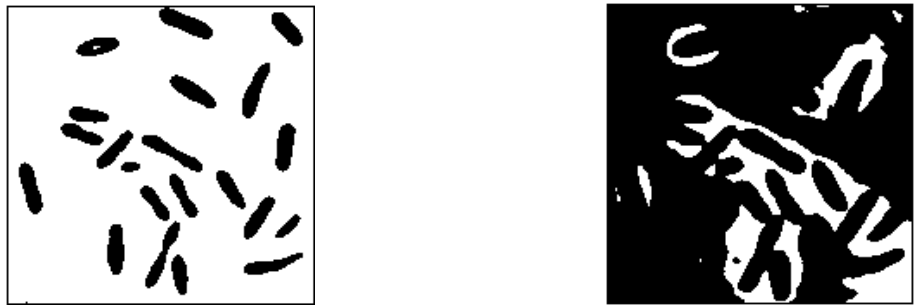

Fig. 3. bacteria image, and its the thresholded images.

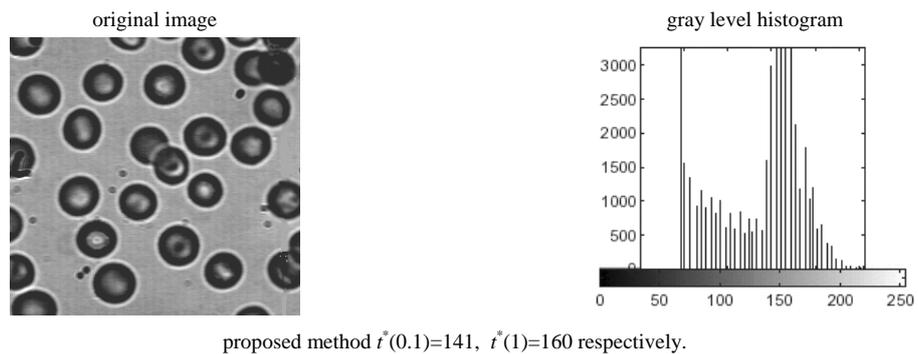

proposed method $t^*(0.1)=141$, $t^*(1)=160$ respectively.

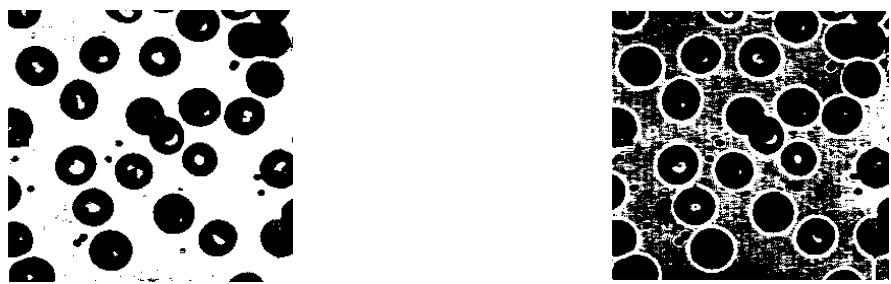

Fig. 4. blood1 image, and its the thresholded images.

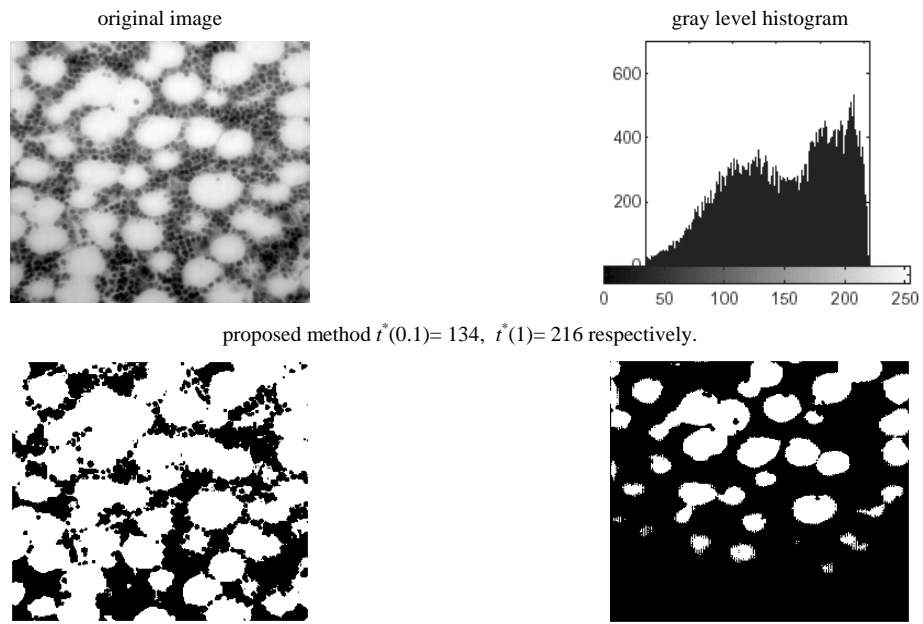

proposed method $t^*(0.1) = 134$, $t^*(1) = 216$ respectively.

Fig. 5. bonemarr image, and its the thresholded images.

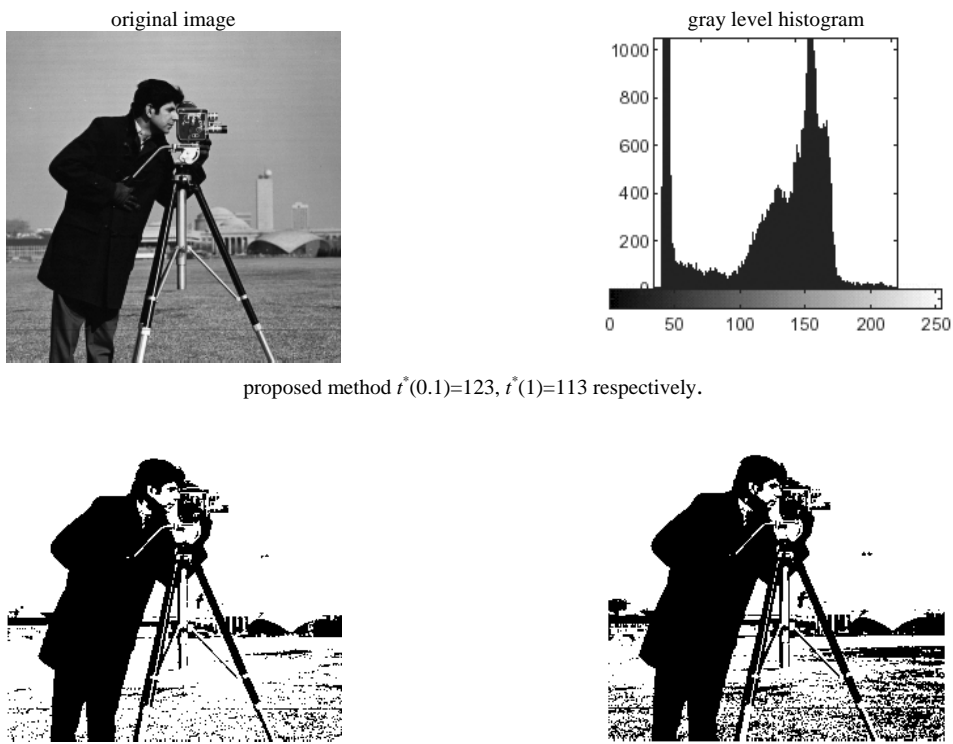

proposed method $t^*(0.1) = 123$, $t^*(1) = 113$ respectively.

Fig. 6. cameraman image, and its the thresholded images.

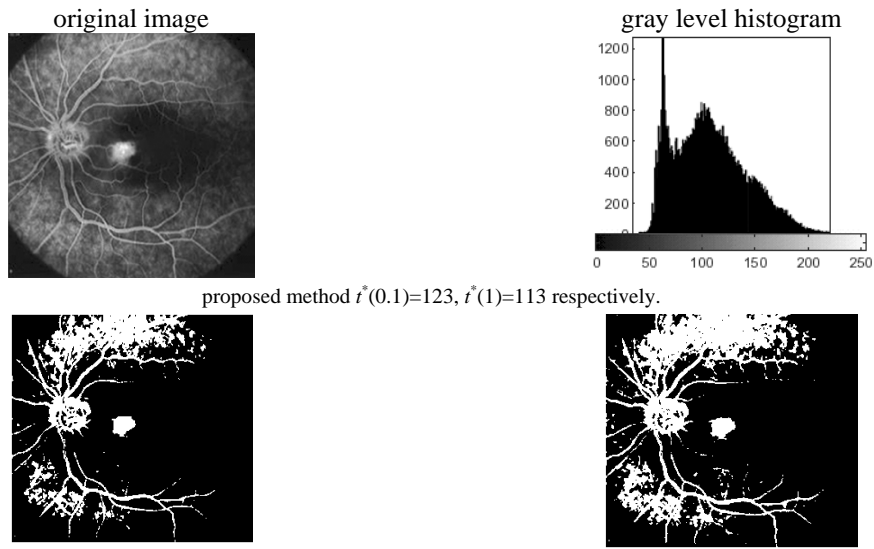

Fig. 7 cell image, and its the thresholded images.

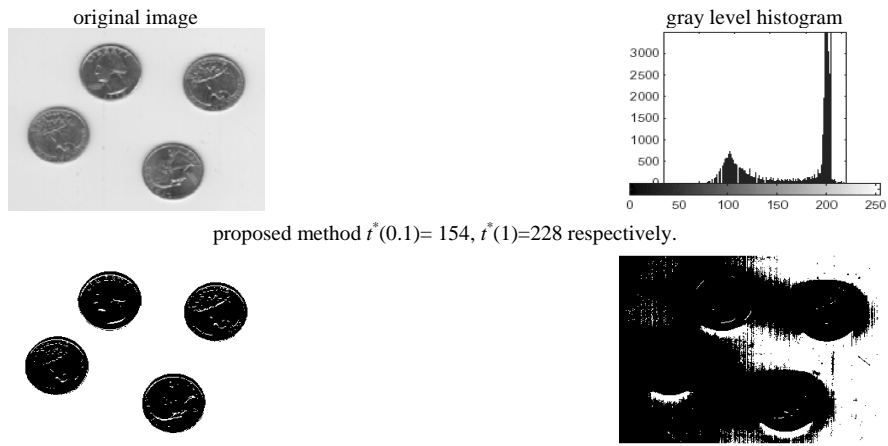

Fig. 3 eight image, and its the thresholded images.

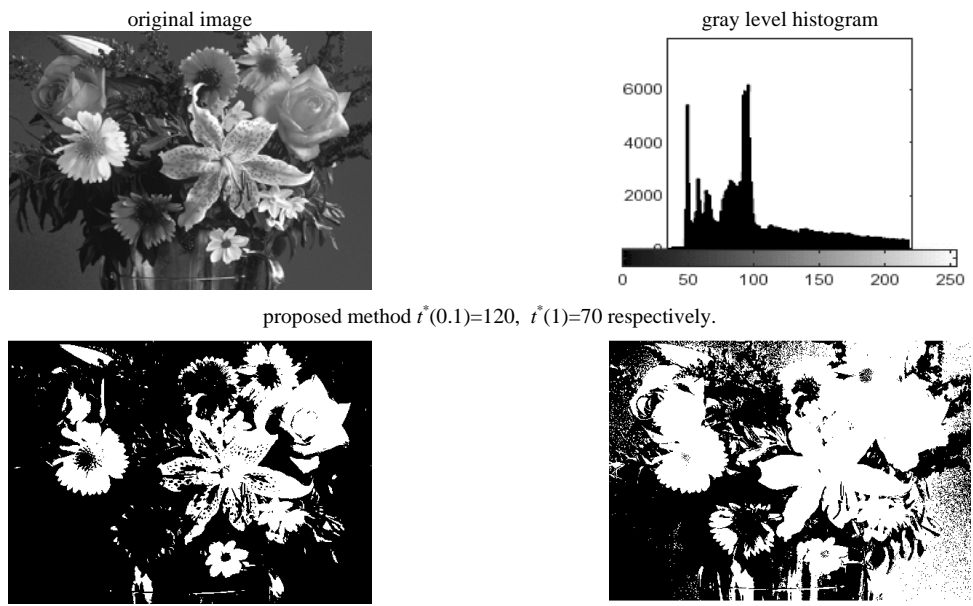

Fig. 9. flowers image, and its the thresholded images.

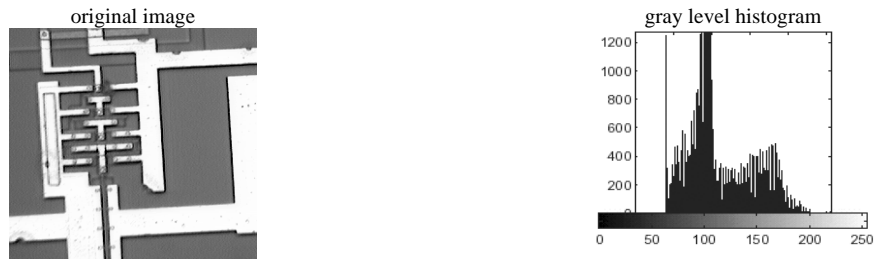

proposed method $t^*(0.1)= 132$, $t^*(1)=23$ respectively.

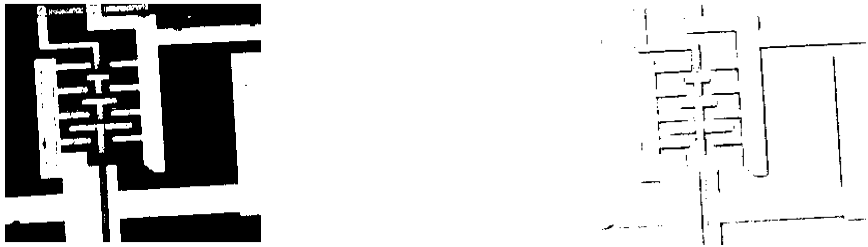

Fig. 10. ic image, and its the thresholded images.

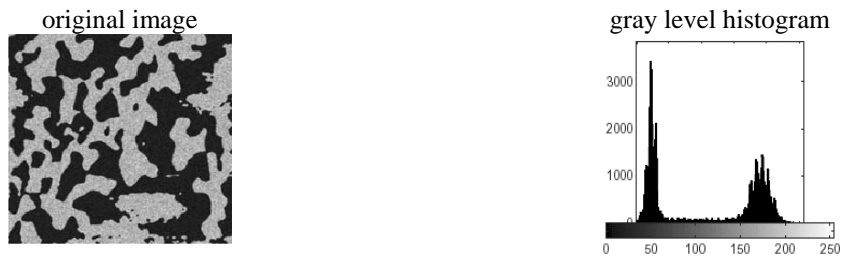

proposed method $t^*(0.1)= 117$, $t^*(1)=151$ respectively.

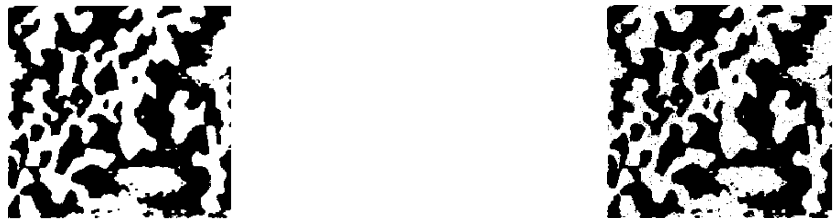

Fig. 11. magnetic image, and its the thresholded images.

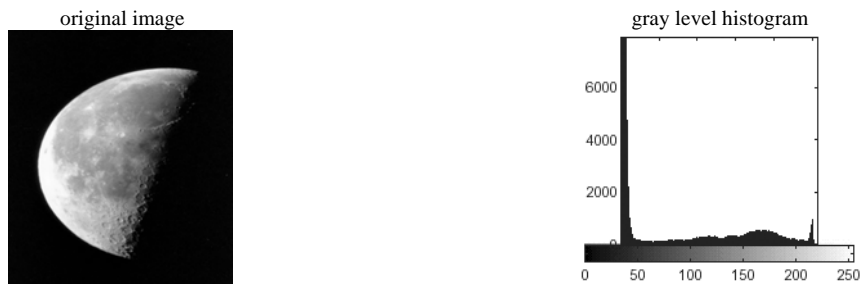

proposed method $t^*(0.1)= 132$, $t^*(1)= 197$ respectively.

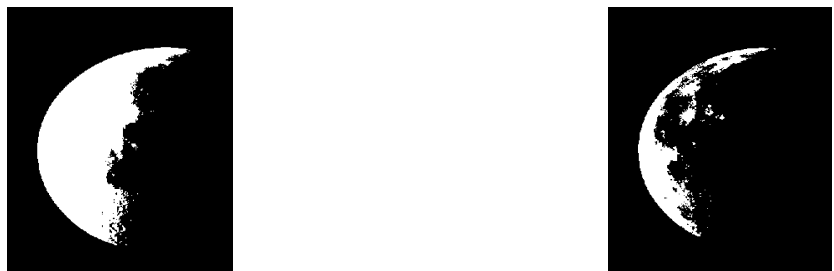

Fig. 12. moon image, and its the thresholded images.

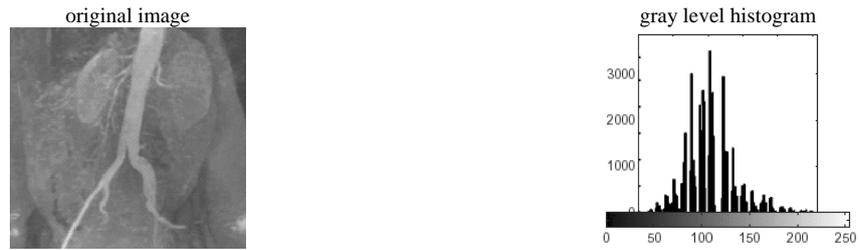

proposed method $t^*(0.1)= 145$, $t^*(1)=132$ respectively.

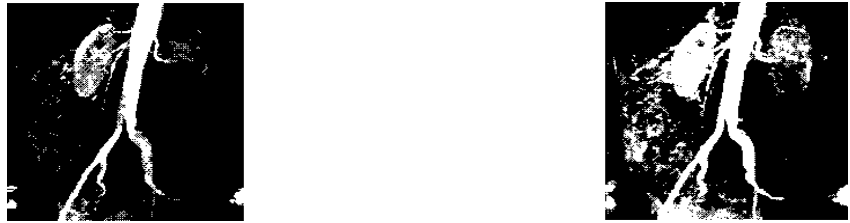

Fig. 13. medical image, and its the thresholded images.

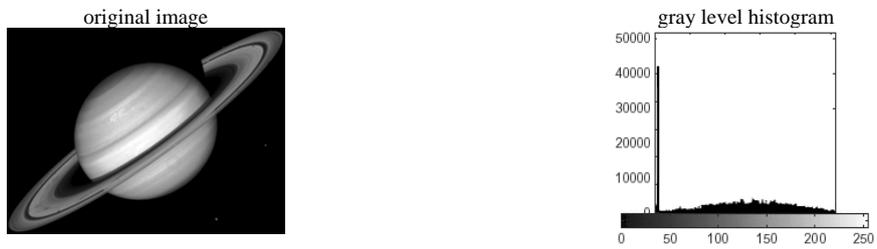

proposed method $t^*(0.1)= 45$, $t^*(1)=165$ respectively.

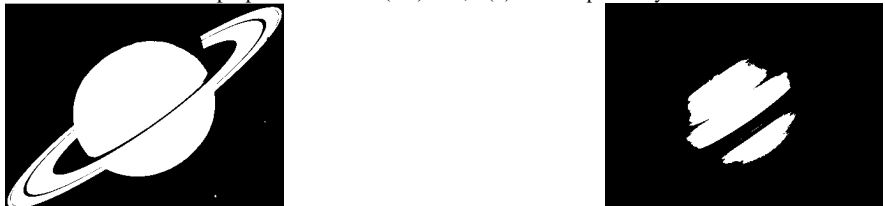

Fig. 14. Saturn image, and its the thresholded images.

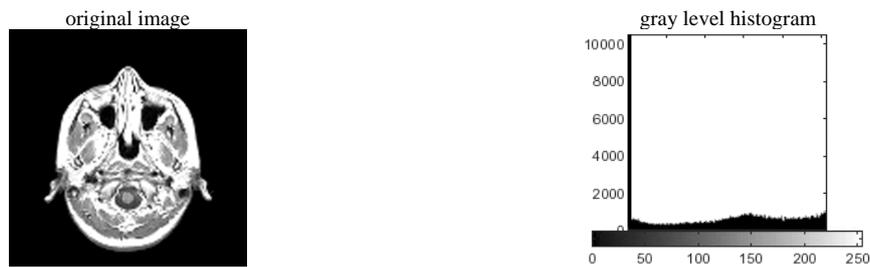

proposed method $t^*(0.1)= 131$, $t^*(1)=101$ respectively.

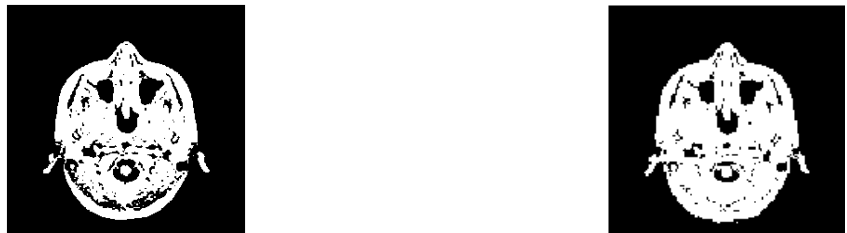

Fig. 15. mri image, and its the thresholded images.

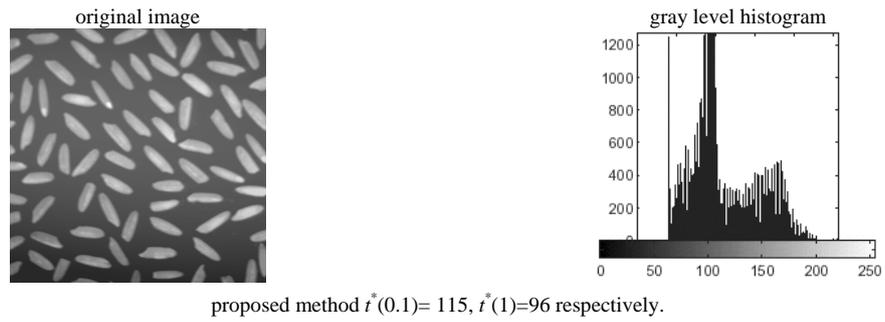

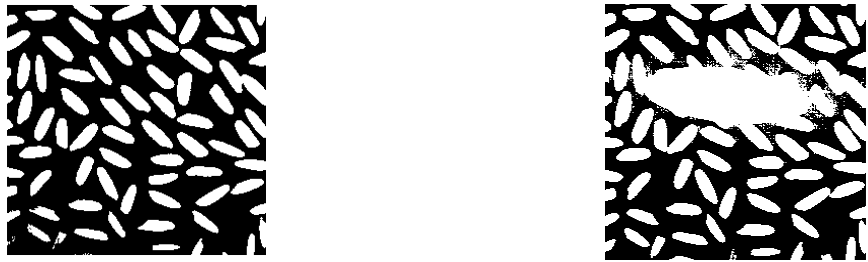

Fig. 16. rice image, and it's the thresholded images.

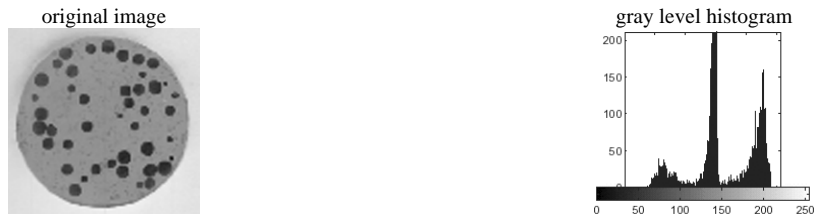

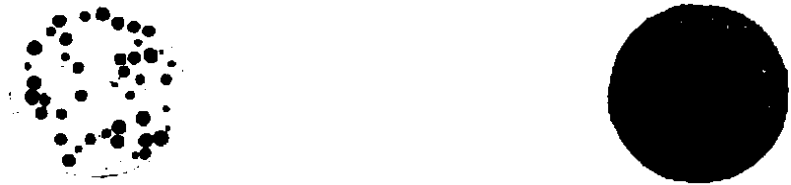

Fig.17. shot1 image, and its the thresholded images.

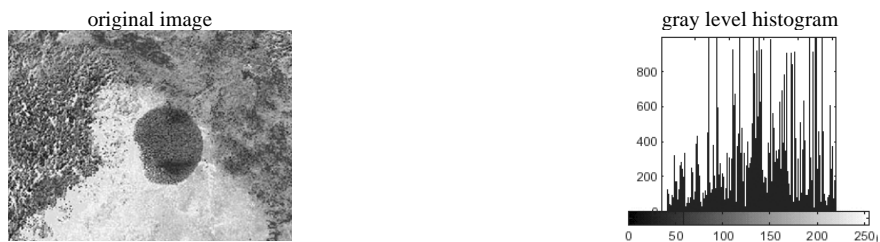

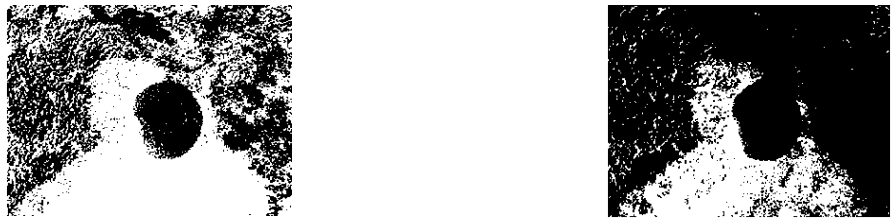

Fig. 18. shadow image, and its the thresholded images.

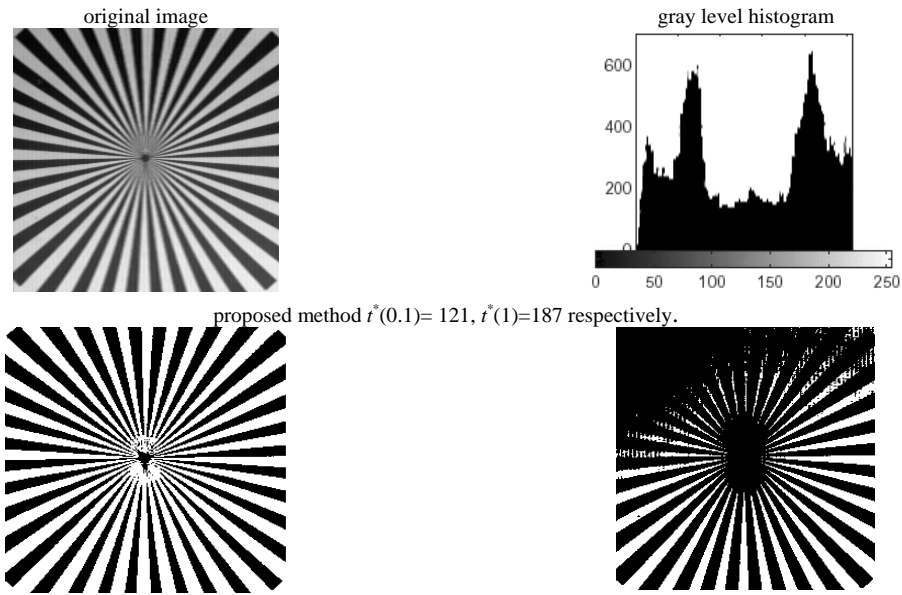

Fig. 19. testpat1 image, and its the thresholded images.

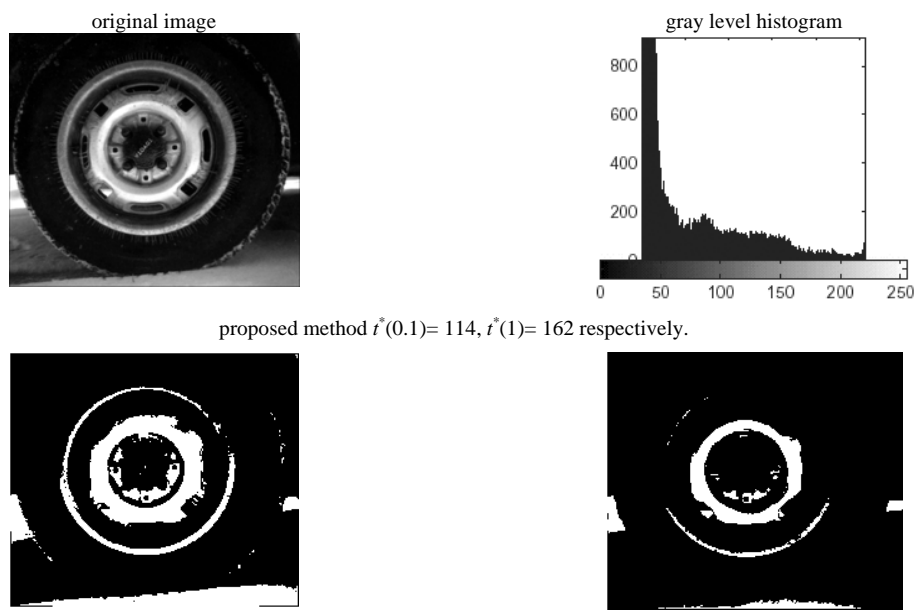

Fig. 20. tire image, and its the thresholded images.

## 5. CONCLUSION

Nonextensive entropy image thresholding is a powerful technique for image segmentation. The presented method has been derived from the generalized entropy concepts proposed by Tsallis of order $q$. The advantage of the method is the use of a global and objective property of the two-dimensional histogram and this method is easily implemented. In almost every image used, the proposed method yielded a good threshold value for fractional $q$ coefficient, that is, when $0<q<1$. For $q>1$, the proposed method did not yield a good threshold value. When $q = 1$, the method yielded, in some cases, good optimal threshold values and, in some other cases, unacceptable threshold values.

The Tsallis $q$ coefficient can be used as an adjustable value and can play an important role as a tuning parameter in the image processing chain for the same class of images. This can be an advantage when the image processing tasks depend on an automatic thresholding.

It is already pointed out in the introduction that the two-dimensional extension gives rise to the exponential increment of computational time. However, the proposed method is decrease the computation time. The software used to generate the results in this paper was written in the commercial software MATLAB 7 on a computer with 2.1GHz Intel Core 2 Duo CPU laptop with 2 GB of RAM, and thus we have taken the advantage of the vector computation that MATLAB offers. Because of this our method takes few seconds.